%% file: main.tex
\documentclass[10pt,twocolumn,letterpaper]{article}

\usepackage[pagenumbers]{cvpr} 

\input{preamble}
\definecolor{cvprblue}{rgb}{0.21,0.49,0.74}
\usepackage[pagebackref,breaklinks,colorlinks,allcolors=cvprblue]{hyperref}
\usepackage{multirow} 

\def\paperID{*****} 
\def\confName{****}
\def\confYear{****}

\newcommand{\btv}{{baller2vec}}
\newcommand{\btvpp}{{baller2vec++}}

\newcommand{\myparagraph}[1]{\noindent{\textbf{#1}~~}}

\title{CourtMotion: Learning Event-Driven Motion Representations from Skeletal Data for Basketball}

\author{Omer Sela\\
Amazon, Tel Aviv University\\
{\tt\small omersela@amazon.com}
\and
Michael Chertok\\
Amazon\\
{\tt\small michael.chertok@gmail.com}
\and
Lior Wolf\\
Tel Aviv University\\
{\tt\small liorwolf@gmail.com}
}

\begin{document}
\maketitle
\input{cvpr/sec/0_abstract}    
\input{cvpr/sec/1_intro}

\input{cvpr/sec/2_related_work}
\input{cvpr/sec/3_method}
\input{cvpr/sec/4_experiments}
\input{cvpr/sec/5_conclusion}

{
    \clearpage
    \small
    \bibliographystyle{ieeenat_fullname}
    \bibliography{main}
}

\clearpage
\input{cvpr/sec/6_appendix}

\end{document}

%% file: cvpr/sec/0_abstract.tex
\begin{abstract}
This paper presents CourtMotion, a spatiotemporal modeling framework for analyzing and predicting game events and plays as they develop in professional basketball. Anticipating basketball events requires understanding both physical motion patterns and their semantic significance in the context of the game. Traditional approaches that use only player positions fail to capture crucial indicators such as body orientation, defensive stance, or shooting preparation motions. Our two-stage approach first processes skeletal tracking data through Graph Neural Networks to capture nuanced motion patterns, then employs a Transformer architecture with specialized attention mechanisms to model player interactions. We introduce event projection heads that explicitly connect player movements to basketball events like passes, shots, and steals, training the model to associate physical motion patterns with their tactical purposes. Experiments on NBA tracking data demonstrate significant improvements over position-only baselines: 35\% reduction in trajectory prediction error compared to state-of-the-art position-based models and consistent performance gains across key basketball analytics tasks. The resulting pretrained model serves as a powerful foundation for multiple downstream tasks, with pick detection, shot taker identification, assist prediction, shot location classification, and shot type recognition demonstrating substantial improvements over existing methods.
\end{abstract}

%% file: cvpr/sec/1_intro.tex
\section{Introduction}
\label{sec:intro}
Basketball is a dynamic team sport where players continuously coordinate their movements to create scoring opportunities and defend against opponents. The complexity of these interactions presents significant technical challenges for automated analysis systems. Anticipating key events and understanding team tactics requires interpreting subtle signals in player movement patterns within the broader game context. These challenges include the coordination of multiple players, the high-dimensional nature of player movements, varying time horizons, and the need to distinguish between routine motion and strategically significant actions.

Recent research has shown promise in addressing these challenges through trajectory prediction pretraining \citep{hauri2022groupactivityrecognitionbasketball, capellera2024hooptransformer}. Models trained to predict player movements develop a robust understanding of game patterns, enabling various capabilities from early shot prediction to play classification. However, these approaches rely solely on player positions, missing crucial signals encoded in player posture and motion that often indicate intent before position changes become apparent.

Our key insight is that effective basketball analysis requires simultaneously understanding motion patterns and their tactical outcomes in the context of the game. While skeletal data captures subtle movements like pivots and accelerations, multi-task training with both trajectory and event prediction losses helps the model connect physical signals to their semantic purpose, distinguishing meaningful patterns from routine motion.

This paper presents CourtMotion, a spatiotemporal model that transforms basketball play analysis through two key innovations: (i) a Graph Neural Network (GNN) that processes high-frequency skeletal joint data, and (ii) event projection attention heads that map player motion embeddings to basketball events occurring in past, present and future windows, forcing the model to learn representations that connect physical movements to their tactical purpose.

The GNN generates rich pose embeddings that capture nuanced movements and subtle signals that often precede and indicate upcoming plays. To complement this, the attention model learns representations that connect motion patterns to their tactical outcomes by simultaneously predicting both player trajectories and fundamental basketball events (passes, shots, rebounds, steals, etc). The joint multi-task optimization, therefore, creates embeddings that encode both physical motion and its strategic importance, making the model particularly effective at understanding complex game dynamics.

Our experiments on NBA tracking data validate this approach, demonstrating a 35\% reduction in trajectory prediction error compared to position-only baselines. More importantly, the model shows strong performance on critical basketball analytics tasks, such as predicting next-shot takers, assists, pick plays, shot location, and shot type classification several seconds before they occur. These capabilities enable advanced analytics applications including automated play recognition, player evaluation metrics, and tactical pattern analysis. The resulting pretrained model serves as a powerful foundation for multiple downstream applications in sports analytics and performance evaluation.

%% file: cvpr/sec/2_related_work.tex
\section{Related Work}
\label{sec:rel_work}
\myparagraph{Multi-Agent Spatiotemporal Modeling}
Multi-agent spatiotemporal modeling (MASM) underpins applications in sports analytics, autonomous driving, and crowd monitoring. Early methods with hand-crafted features \citep{kim2010motion} evolved into deep learning approaches: \citet{zheng2016generating} used hierarchical neural nets for trajectory forecasting, while  \citet{hauri2022groupactivityrecognitionbasketball} employed conditional VAEs for adversarial motion prediction. Graph-based architectures naturally represent inter-agent relationships \citep{yeh2019diverse,ivanovic2018generative}, though they struggled with preserving individual characteristics at scale. For skeletal motion specifically,  \citet{Shi_2019_CVPR} introduced directed graph neural networks (DGNNs) that leverage joint connectivity and directionality, providing a foundation for effectively processing human pose data in action recognition tasks.

\myparagraph{Transformers for Spatiotemporal Data}
Transformers \citep{vaswani2017attention}, initially from NLP, have revolutionized sequence modeling across domains \citep{dosovitskiy2021image}. For trajectory forecasting, \citet{giuliari2020transformer} adapted transformers to pedestrian paths. A significant advancement came with \btv{} ~\citep{alcorn2021baller2vec}, which addressed multi-agent spatiotemporal systems but maintained an assumption that agent trajectories were statistically independent at each time step- a fundamental limitation for modeling coordinated behaviors.

\myparagraph{Modeling Coordinated Agent Behavior}
The challenge of modeling coordinated behavior without statistical independence assumptions has been addressed through various approaches. \citet{zhan2019generating} introduced shared "macro-intents" influencing multiple agents, while \btvpp{}~\citep{alcorn2021baller2vec++} rethought information flow in multi-entity transformers through look-ahead trajectories and masking strategies. Recent sports-specific architectures include FootBots \citep{capellera2024footbots} for soccer motion prediction and TranSPORTmer \citep{capellera2024transportmer} for multi-task trajectory understanding in team sports.

%% file: cvpr/sec/3_method.tex
\section{Method}
\label{sec:method}
\begin{figure*}[htbp]
 \centering
 \includegraphics[width=\linewidth]{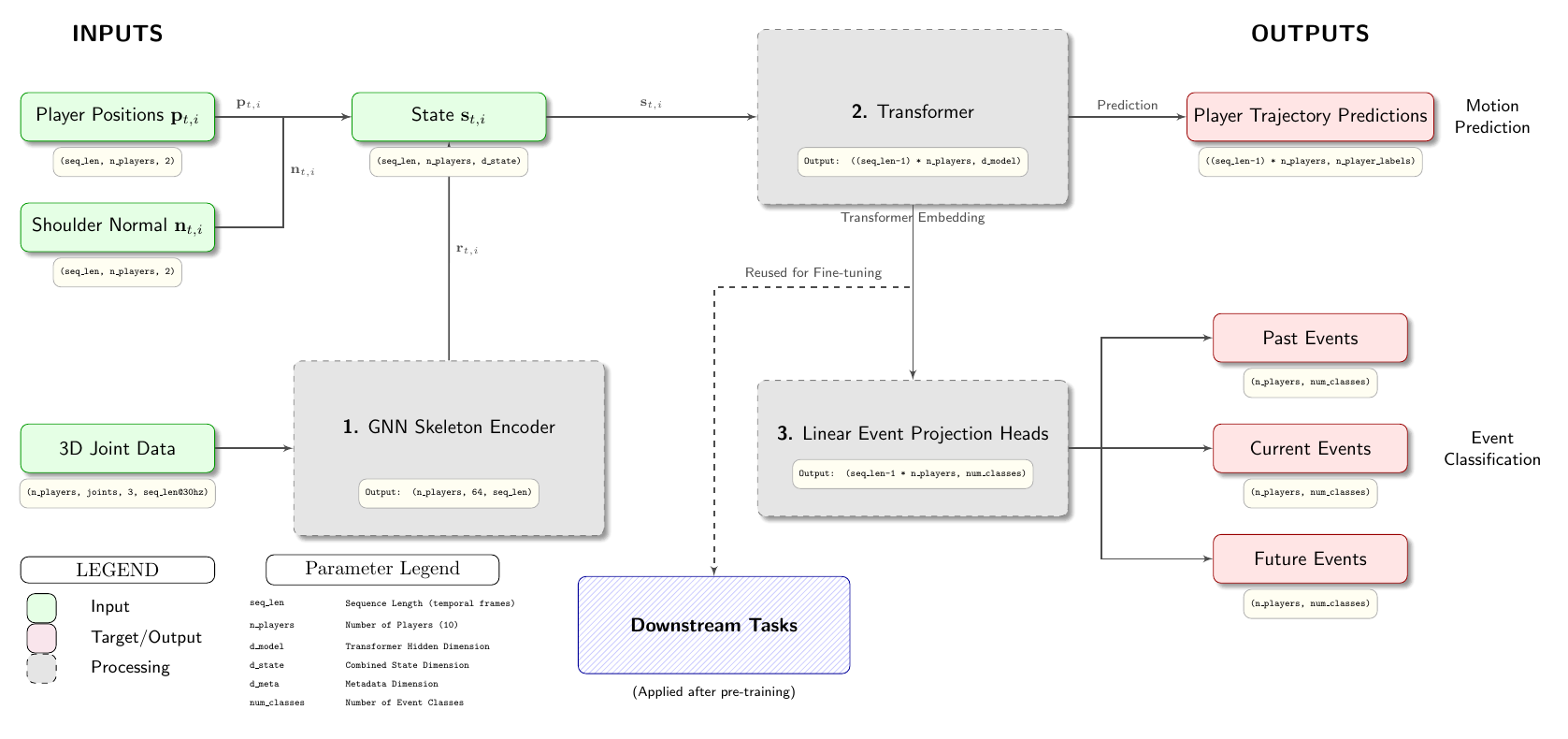}
 \caption{CourtMotion architecture with three main components: (1) GNN Skeleton Encoder that processes 3D joint data into player motion embeddings, (2) Transformer-based trajectory predictor integrating motion with position data, and (3) Event Projection Heads that predict basketball events.}
 \label{fig:app_diagram}
\end{figure*}
Consider a set of $N$ basketball players, indexed by $i \in \{1, 2, ..., N\}$, moving on a court over discrete time steps $t \in \{1, 2, ..., T\}$. At each time step $t$, the position of player $i$ is represented by a 2D coordinate $\mathbf{p}_{t,i} = (x_{t,i}, y_{t,i})$ sampled at 5 Hz.
Let $\ell = \{\text{no\_action}, \text{free\_throw}, \text{shot\_attempt}, \text{pass}, \text{deflection}, \text{block}, \\ \text{rebound}, \text{steal}, \text{dribble}\}$ be the set of basketball events. At each time step $t$, the binary indicator of event $\varphi \in \ell$ for player $i$ is denoted as $\epsilon_{t,i}^{\varphi} \in \{0, 1\}$.
Our pretraining objective is twofold: 
\begin{enumerate}[leftmargin=15pt]
    \item To predict the position change delta: $\Delta \mathbf{p}_{t,i} = \mathbf{p}_{t+1,i} - \mathbf{p}_{t,i}$ \item To predict events across three temporal regions relative to the current timestep: \begin{align}
\epsilon_{t,i,\text{past}}^{\varphi} &= \max_{\tau \in [-\delta_{\text{past}}, -1]} \epsilon_{t+\tau,i}^{\varphi} \\
\epsilon_{t,i,\text{current}}^{\varphi} &= \epsilon_{t,i}^{\varphi} \\
\epsilon_{t,i,\text{future}}^{\varphi} &= \max_{\tau \in [1, \delta_{\text{future}}]} \epsilon_{t+\tau,i}^{\varphi}
\end{align}
\end{enumerate}
where $\epsilon_{t,i,\text{past}}^{\varphi}$, $\epsilon_{t,i,\text{current}}^{\varphi}$, and $\epsilon_{t,i,\text{future}}^{\varphi}$  represent the event occurrences for past, current, and future windows, respectively. Parameters $\delta_{\text{past}} > 0$ and $\delta_{\text{future}} > 0$ define the extent of these windows. During training, we employ linear classification heads to predict, based on the player embeddings, events in the three temporal windows, enabling the model to learn which motion patterns precede, coincide with, or follow different basketball actions. These projections are crucial for developing representations that connect physical movements to their tactical significance in the context of the game.
In addition to positions $\mathbf{p}_{t,i}$, each player's state $\mathbf{s}_{t,i}$ contains two components: The first component is the \textbf{Pose Embedding $\mathbf{r}_{t,i}$}, composed of 30 Hz 3D skeletal joints compressed to 5 Hz via our GNN. The second additional component is the \textbf{Shoulder-Normal Vector $\mathbf{n}_{t,i}$}: Given left and right shoulder joint coordinates $\mathbf{u}_{t,i}^{(L)}$ and $\mathbf{u}_{t,i}^{(R)}$:\begin{equation}
\begin{array}{cc}
   \mathbf{w}_{t,i} = \mathbf{u}_{t,i}^{(R)} - \mathbf{u}_{t,i}^{(L)} &
   \mathbf{n}_{t,i} = \frac{1}{\|\mathbf{w}_{t,i}\|} \begin{pmatrix} 0 & -1\\ 1 & \phantom{-}0 \end{pmatrix} \mathbf{w}_{t,i}
\end{array}
\end{equation} 
The shoulder-normal vector serves as a computationally lightweight alternative to full skeletal tracking, providing crucial directional information about player orientation with minimal overhead. We explicitly include it separately from the full pose embedding to evaluate its independent contribution, as orientation data can be extracted from broadcast footage or basic tracking systems when comprehensive skeletal data is unavailable. This separation enables practical deployment scenarios where computational or sensor constraints may limit access to complete biomechanical features while still capturing directional intent, a critical signal for anticipating player movements and actions. 
The complete player state and its delta are defined as:
\begin{equation}
\begin{array}{cc}
   \mathbf{s}_{t,i}=(\mathbf{r}_{t,i}, \mathbf{n}_{t,i}, \mathbf{p}_{t,i}) &
   \Delta \mathbf{s}_{t,i} = \mathbf{s}_{t+1,i} - \mathbf{s}_{t,i}
\end{array}
\end{equation}

Our architecture consists of three key components depicted in Figure \ref{fig:app_diagram}: A GNN Skeleton Encoder, a Transformer, and linear projection heads to classify the events.

\subsection{GNN Skeleton Encoder} 
\begin{figure*}[htbp]
 \centering
 \includegraphics[width=0.68\linewidth]{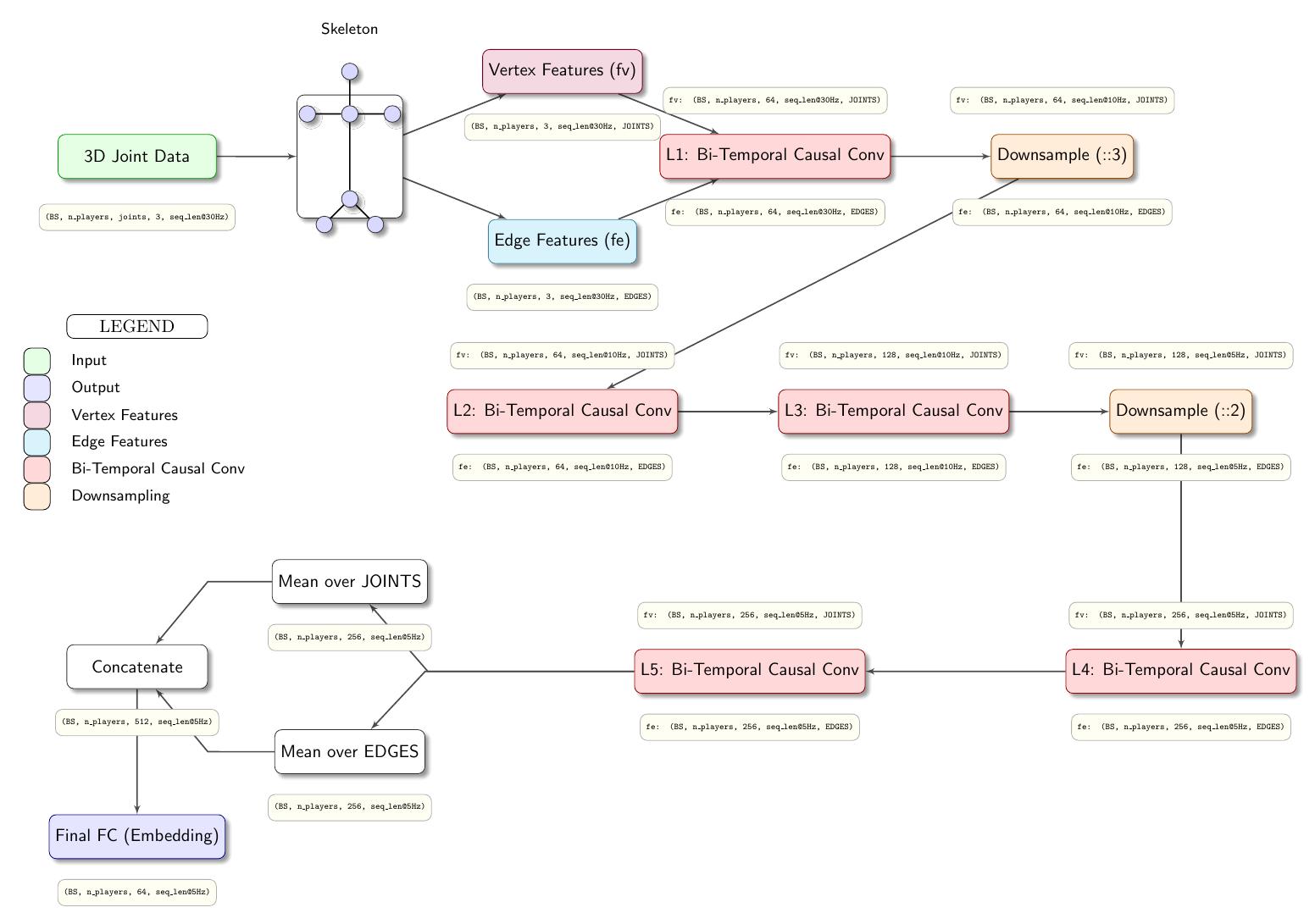}
 \caption{GNN Skeleton Encoder processes 3D joint data through parallel vertex and edge processing streams with bi-temporal causal convolutions and strategic downsampling.}
 \label{fig:gnn_encoder}
\end{figure*}
This section details how we process player skeletons at 30 Hz to produce 5 Hz pose embeddings using a directed graph neural network approach. Figure \ref{fig:gnn_encoder}  shows the structure of our GNN encoder.
\myparagraph{Skeleton as a Directed Acyclic Graph}
At 30 Hz, each player's skeleton is defined by a set of joints $\mathcal{V}$ and directed edges $\mathcal{E}$. Let $\mathbf{v}_{\tau,i}^{(j)} = (x_{\tau,i}^{(j)},\,y_{\tau,i}^{(j)},\,z_{\tau,i}^{(j)})$ be the 3D coordinates of joint $j\in\mathcal{V}$ for player $i$ at time $\tau$. We define the root as the middle hip joint, with edges pointing outward from more central joints to more peripheral ones. Formally, each directed edge $(m \to j)\in\mathcal{E}$ is associated with a 3D bone vector:

\begin{equation}\mathbf{b}_{\tau,i}^{(m\to j)} = \mathbf{v}_{\tau,i}^{(m)} \;-\; \mathbf{v}_{\tau,i}^{(j)}.\end{equation}

Hence, each skeleton frame $\tau$ forms a directed acyclic graph (DAG) whose vertices carry 3D joint positions, while edges carry 3D bone vectors.

\myparagraph{DGN Block for Vertex and Edge Updates}
A DGN block \citep{Shi_2019_CVPR} updates all vertices and edges in the DAG by aggregating information from incoming/outgoing edges and neighboring vertices. This process relies on two key operations, one to update vertices, and one to update edges, along with aggregator functions that handle multiple edges connected to a single vertex.

Let us denote the feature of joint $j$ at layer $\ell$ by $\mathbf{h}_{\tau,i}^{(j,\ell)}$, and the feature of an edge $(m \to j)$ by $\mathbf{e}_{\tau,i}^{(m\to j,\ell)}$. At the input layer $\ell=0$, we initialize:
\begin{equation}\mathbf{h}_{\tau,i}^{(j,0)} \;=\; \mathbf{v}_{\tau,i}^{(j)}, \quad \mathbf{e}_{\tau,i}^{(m\to j,0)} \;=\; \mathbf{b}_{\tau,i}^{(m\to j)}.\end{equation}
\myparagraph{Aggregation}
For vertex $j$, let $\mathcal{E}_{j}^{-}$ be the set of edges that point into $j$, and $\mathcal{E}_{j}^{+}$ be the set of edges that point out of $j$. We define the summation aggregation functions, $g_{e}^{-}$ and $g_{e}^{+}$, which take all incoming or outgoing edges and produce a single "aggregated" feature by summing all edges provided to them:
\begin{equation}
\begin{aligned}
\overline{\mathbf{e}}_{j}^{-} &= g_{e}^{-}\bigl(\{\mathbf{e}_{\tau,i}^{(e,\ell)} : e\in \mathcal{E}_{j}^{-}\}\bigr) = \sum_{e \in \mathcal{E}_{j}^{-}} \mathbf{e}_{\tau,i}^{(e,\ell)}, \\
\overline{\mathbf{e}}_{j}^{+} &= g_{e}^{+}\bigl(\{\mathbf{e}_{\tau,i}^{(e,\ell)} : e\in \mathcal{E}_{j}^{+}\}\bigr) = \sum_{e \in \mathcal{E}_{j}^{+}} \mathbf{e}_{\tau,i}^{(e,\ell)}.
\end{aligned}
\end{equation}
By summing the features in each set, we obtain two fixed-length vectors summarizing all edges into and out of vertex $j$.
\myparagraph{Vertex Update}
For the vertex update of $\mathbf{h}_{\tau,i}^{(j,\ell)}$ we concatenate it with the aggregated incoming/outgoing edge features, $\overline{\mathbf{e}}_{j}^{-}$ and $\overline{\mathbf{e}}_{j}^{+}$, and pass through an update function $h_v$:   
\begin{equation}\mathbf{h}_{\tau,i}^{(j,\ell+1)} \;=\; h_{v}\!\Bigl(\bigl[\mathbf{h}_{\tau,i}^{(j,\ell)},\,\overline{\mathbf{e}}_{j}^{-},\,\overline{\mathbf{e}}_{j}^{+}\bigr]\Bigr).\end{equation}
In practice, $h_v$ is a single fully connected (FC) layer followed by a BatchNorm and a non-linear activation ReLU.
\myparagraph{Edge Update}
After the vertices have been updated, each edge $(m \to j)$ is revisited. We concatenate the edge's current feature $\mathbf{e}_{\tau,i}^{(m\to j,\ell)}$ with the newly updated source and target vertices $\mathbf{h}_{\tau,i}^{(m,\ell+1)}$ and $\mathbf{h}_{\tau,i}^{(j,\ell+1)}$. This combined vector is fed into another FC-based update function $h_e$:
\begin{equation}\mathbf{e}_{\tau,i}^{(m\to j,\ell+1)} \;=\; h_{e}\!\Bigl( \bigl[ \mathbf{e}_{\tau,i}^{(m\to j,\ell)},\; \mathbf{h}_{\tau,i}^{(m,\ell+1)},\; \mathbf{h}_{\tau,i}^{(j,\ell+1)} \bigr] \Bigr).\end{equation}
In practice, $h_e$ is a single fully connected (FC) layer followed by a BatchNorm and a non-linear activation ReLU.
\myparagraph{BiTemporal Causal Convolutions with Striding}
After each DGN block processes the spatial information within a frame $\tau$, we use a shared {BiTemporal Causal Convolution} to model temporal dynamics. We apply the same causal 1D convolutional operator to both vertex features $\mathbf{h}_{\tau,i}^{(j,\ell)}$ and edge features $\mathbf{e}_{\tau,i}^{(m\to j,\ell)}$. Let $\mathbf{z}_{\tau,i}^{(\ell)}\in \mathbb{R}^{d}$ represent either vertex or edge states at layer $\ell$. A causal convolution of kernel size $K$ and stride $s$ updates:
\begin{equation}
\label{eq:temporal_conv}
\widetilde{\mathbf{z}}_{\tau,i}^{(\ell)}
\;=\;
\sigma\!\Bigl(
  \sum_{k=0}^{K-1}
    \mathbf{W}_k \,\mathbf{z}_{(\tau - k),\,i}^{(\ell)}
  \;+\;
  \mathbf{b}
\Bigr),
\end{equation}
where $\tau$ runs over time indices, $\mathbf{W}_k \in \mathbb{R}^{d\times d}$ are learnable weights, $\mathbf{b}\in\mathbb{R}^{d}$, and $\sigma(\cdot)$ is a non-linear activation. This convolution is \emph{causal}: it only sums over current and previous frames $\tau'\le \tau$.
\myparagraph{DGNN Pipeline}
Concretely, one block of our DGNN pipeline consists of:
(i) DGN Block (vertex update, edge update) to aggregate spatial relationships within frame $\tau$, 
(ii) BiTemporal Causal Convolution to aggregate motion across consecutive frames up to $\tau$, and 
(iii) Temporal stride downsampling to reduce from 30Hz to 5Hz.
By stacking these blocks multiple times, the model produces multi-scale spatiotemporal features for each joint and edge. The final pose embedding $\mathbf{r}_{t,i}$ is obtained by aggregating across nodes and edges, concatenating, and applying a final fully connected layer.

\subsection{Player Embedding  Transformer} 

We extend the Baller2Vec++ \citep{alcorn2021baller2vec++} architecture to incorporate skeletal data and utilize event predictions. For each player $i$ at time step $t$, we define three feature vectors: 1. Current state including skeleton embedding and shoulder normal direction: $\mathbf{z}_{t,i} = g_z([\boldsymbol{\psi}(i), \mathbf{p}_{t,i}, \mathbf{s}_{t,i}])$ 2. Look-ahead trajectory and state: $\mathbf{u}_{t,i} = g_u([\boldsymbol{\psi}(i), \mathbf{p}_{t+1,i}, \mathbf{s}_{t,i}, \Delta \mathbf{p}_{t,i}, \Delta \mathbf{s}_{t,i}])$ 3. Starting location and initial state: $\mathbf{r}_{0,i} = g_r([\boldsymbol{\psi}(i), \mathbf{p}_{1,i}, \mathbf{s}_{1,i}])$
where $g_z$, $g_u$, and $g_r$ are MLPs, and $\boldsymbol{\psi}(i)$ is a learnable player embedding. These vectors are arranged in input matrix $\mathbf{Z} \in \mathbb{R}^{(2TN + N) \times F}$ and processed by a transformer with an attention mask ensuring that:
\begin{enumerate}[leftmargin=15pt]
\item $\mathbf{r}_{0,i}$ can attend to all $\mathbf{r}_{0,j}$
\item $\mathbf{z}_{t,i}$ can attend to all $\mathbf{r}_{0,j}$, any $\mathbf{z}_{t',j}$ where $t' < t$ or $t' = t$ and $j \leq i$, and any $\mathbf{u}_{t',j}$ where $t' < t$ or $t' = t$ and $j < i$
\item $\mathbf{u}_{t,i}$ can attend to all $\mathbf{r}_{0,j}$, any $\mathbf{z}_{t',j}$ where $t' < t$ or $t' = t$ and $j \leq i$, and any $\mathbf{u}_{t',j}$ where $t' < t$ or $t' = t$ and $j \leq i$
\end{enumerate}
This masking strategy ensures temporal causality while allowing dependency modeling between players. The transformer processes these inputs to produce output embeddings $\hat{\mathbf{Z}}$, from which we extract the transformed representations $\hat{\mathbf{z}}_{t,i}$ for each player and timestep.

\myparagraph{Event Projection Heads}
For each transformer output embedding $\hat{\mathbf{z}}_{t,i}$, representing a player's state, we project it onto basketball events across three temporal windows using linear heads. This projection maps motion patterns to their tactical outcomes, helping the model learn which movements signal different events:
\begin{equation}
\hat{\boldsymbol{\epsilon}}_{t,i,r} = \sigma(W_r \hat{\mathbf{z}}_{t,i} + b_r), \quad r \in \{\text{past}, \text{current}, \text{future}\}
\end{equation}
where $W_r \in \mathbb{R}^{|\ell| \times d}$ and $b_r \in \mathbb{R}^{|\ell|}$ are learnable parameters, $\sigma$ is the sigmoid activation, and $\hat{\boldsymbol{\epsilon}}_{t,i,r} \in \mathbb{R}^{|\ell|}$ outputs probabilities for each event type.

The use of three temporal windows (past, current, future) allows the model to capture the full context of basketball events. The `past' window helps identify precursor movements, the `current' window focuses on immediate actions, and the `future' window encourages the model to anticipate upcoming events. This multi-window approach enables CourtMotion to understand the complete lifecycle of basketball plays, from setup to execution to follow-through.

\myparagraph{Training Objective}
For trajectory prediction, movements are discretized into an 11ft×11ft grid, yielding 121 trajectory bins. Let $\mathcal{B}$ be the set of bins and $v_{t,i} = \text{Bin}(\Delta \mathbf{p}_{t,i}) \in \mathcal{B}$ be the ground truth bin for player $i$ at time $t$. The prediction probability is:
\begin{equation}
P(v_{t,i} | \hat{\mathbf{Z}}) = \text{softmax}(W_{\text{traj}} \cdot \hat{\mathbf{z}}_{t,i} + b_{\text{traj}})
\end{equation}
where $W_{\text{traj}} \in \mathbb{R}^{|\mathcal{B}| \times d}$ and $b_{\text{traj}} \in \mathbb{R}^{|\mathcal{B}|}$ are learnable parameters.
The trajectory loss is the negative log-likelihood:
\begin{equation}
L_{\text{traj}} = \frac{1}{TN}\sum_{t=1}^{T} \sum_{i=1}^{N} -\ln(P(v_{t,i} | \hat{\mathbf{Z}}))
\end{equation}
For event prediction, we use binary cross-entropy:
\begin{equation}
L_{\text{events}} = \frac{1}{TN} \sum_{t=1}^{T} \sum_{i=1}^{N} \sum_{\substack{r \in \{\text{past},\\ \text{current},\\ \text{future}\}}} \sum_{\varphi \in \ell} \text{BCE}(\hat{\epsilon}_{t,i,r}^{\varphi}, \epsilon_{t,i,r}^{\varphi})
\end{equation}
where BCE denotes binary cross-entropy: $\text{BCE}(p,y) = -y\log(p) - (1-y)\log(1-p)$.
The total loss combines both objectives with a weighting parameter $\alpha \in [0,1]$:
\begin{equation}
L_{\text{total}} = \alpha \cdot L_{\text{traj}} + (1-\alpha) \cdot L_{\text{events}}
\end{equation}

%% file: cvpr/sec/4_experiments.tex
\section{Experiments}
\label{sec:experiments}
\myparagraph{Implementation details} 
Our model uses 6 transformer layers with 8 attention heads and hidden dimension of 512. The GNN encoder consists of 5 DGNN blocks. For event projection, we set both temporal window parameters $\delta_{\text{past}}$ and $\delta_{\text{future}}$ to 2 seconds each. The combined loss function uses $\alpha = 0.5$, equally weighting trajectory prediction and event projection objectives. Training used Adam optimizer with learning rate 5e-4 and batch size 32 on 8 NVIDIA A100 GPUs for approximately 100 hours. 
\myparagraph{Dataset} Our experiments utilize the NBA SportVU dataset, containing spatiotemporal recordings from 631 games during the 2015-2016 season. A key innovation in our approach is augmenting this dataset with full skeletal pose information, enriching the standard player centroid positions (x,y coordinates) with joint positions and kinematic relationships. This enhancement enables our model to capture subtle posture changes and directional intent that is not discernible from positions alone. Unlike previous approaches that analyze isolated possession sequences, we process continuous gameplay segments regardless of possession changes, better reflecting real gameplay dynamics while requiring more sophisticated modeling of team coordination patterns. For experiments, we used 160 games for pretraining, 160 different games for training downstream classifiers, and 40 games for evaluation of both stages.

\subsection{Pretraining Results}
\begin{figure*}[htbp]
\centering
\includegraphics[width=0.55\linewidth]{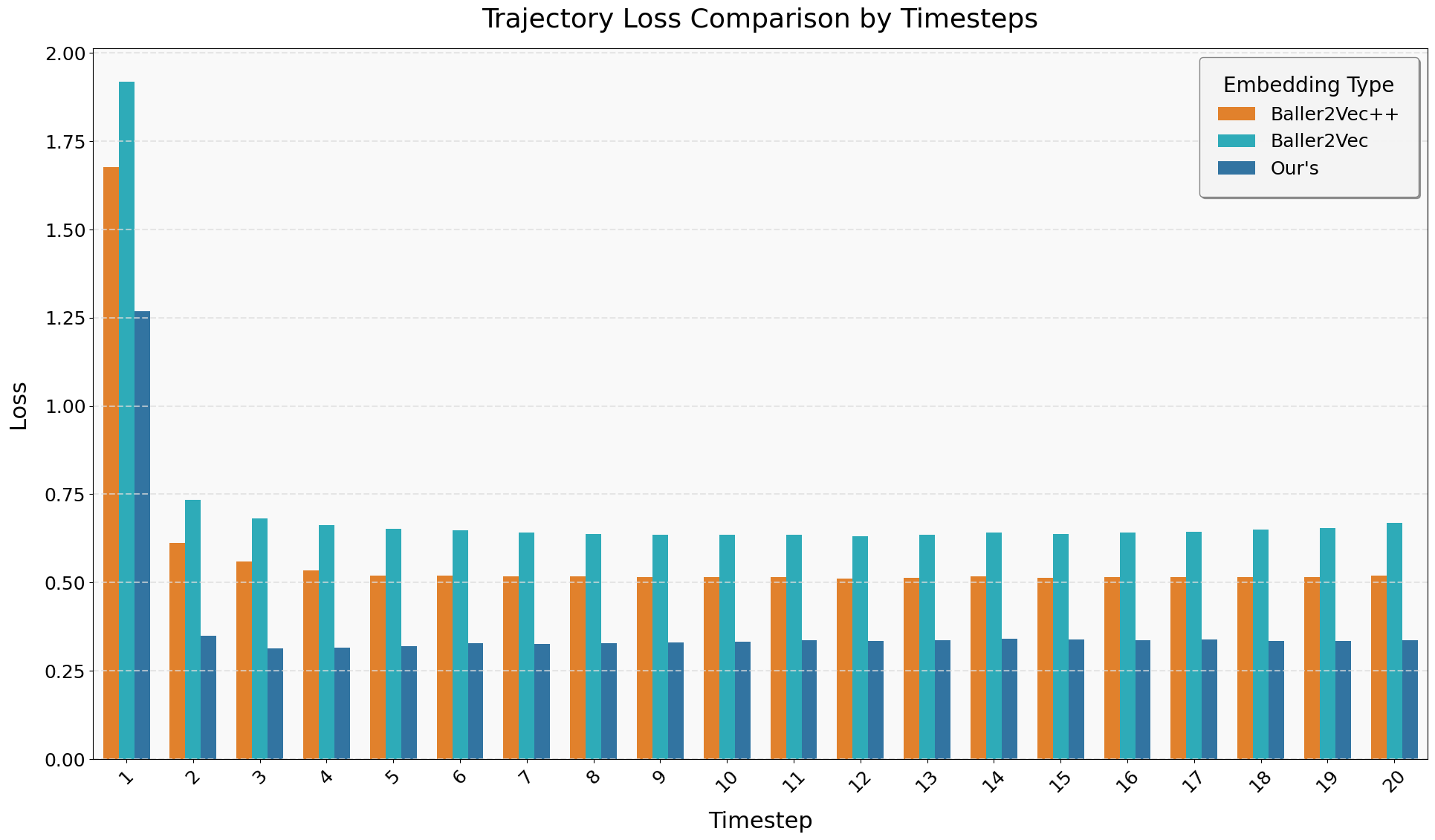}
\caption{Trajectory loss comparison across models at different timesteps, showing consistent performance advantages of our approach over previous methods throughout the sequence.}
\label{fig:loss_by_timestep_compact}
\end{figure*}

\
Following the evaluation protocol established by Baller2Vec++ \citep{alcorn2021baller2vec++}, we evaluate trajectory prediction performance using negative log-likelihood (NLL) over our test set. Our approach achieves a trajectory NLL of 0.38, representing a 35\% improvement over our implementation of Baller2Vec++ (0.58) and a 47\% improvement over the original Baller2Vec (0.71). These substantial gains highlight the effectiveness of our motion-aware representation in capturing the complex dynamics of player movement. Interestingly, our models with and without events projection achieve identical trajectory NLL (0.38), suggesting that while event projection enriches the representation for downstream tasks (as detailed in the next sections), it does not help the task of trajectory prediction. Figure \ref{fig:loss_by_timestep_compact}  shows the consistent advantage of our approach across all timesteps in the prediction sequence.

\subsection{Downstream Tasks}
We evaluate CourtMotion on key coordinated plays using a standardized linear probing methodology: (i) Extract embeddings $\hat{\mathbf{z}}_{t,i}$ from each model variant at every timestamp, (ii) Train a simple linear classifier $clf(\hat{\mathbf{z}}_{t,i}) = \mathbf{W} \hat{\mathbf{z}}_{t,i} + \mathbf{b}$ on these embeddings, (iii) Evaluate at multiple time horizons before events occur, and (iv) Compare against strong baselines including a Baller2Vec++ backbone trained end-to-end without pretraining, and TranSPORTmer \citep{capellera2024transportmer}, a recent multi-agent sports trajectory model. For TranSPORTmer, we extract player embeddings from their pretrained model and apply the same linear probing protocol, ensuring fair comparison across all methods.

\begin{figure}[htbp]
\centering
\centering
\includegraphics[width=\linewidth]{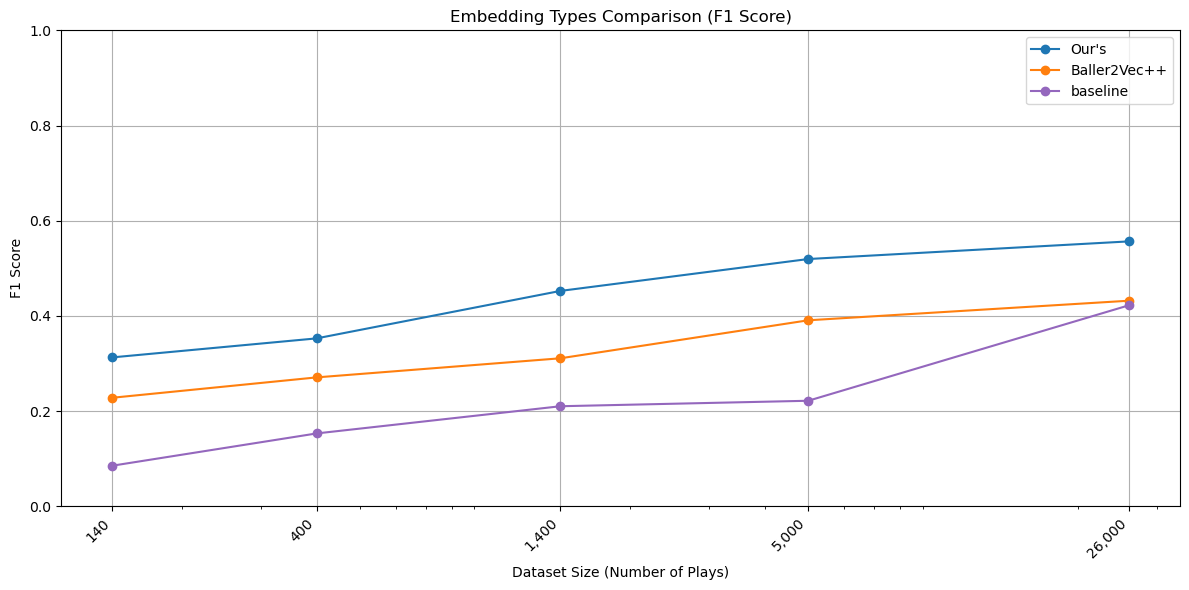}
\caption{Comparison for shot taker prediction across  training set sizes.}
\label{fig:shot_embedding_comparison}
\end{figure}

\myparagraph{Shot Taker Prediction Task}
The shot taker prediction task challenges models to identify which player will take the next shot- a complex anticipatory task that requires understanding of both individual intent and team dynamics. Successful prediction demands recognition of subtle movement patterns that signal shooting preparation as players position themselves for scoring opportunities. 

\begin{figure*}[t]
\centering
\begin{subfigure}[b]{0.48\textwidth}
    \includegraphics[width=\textwidth]{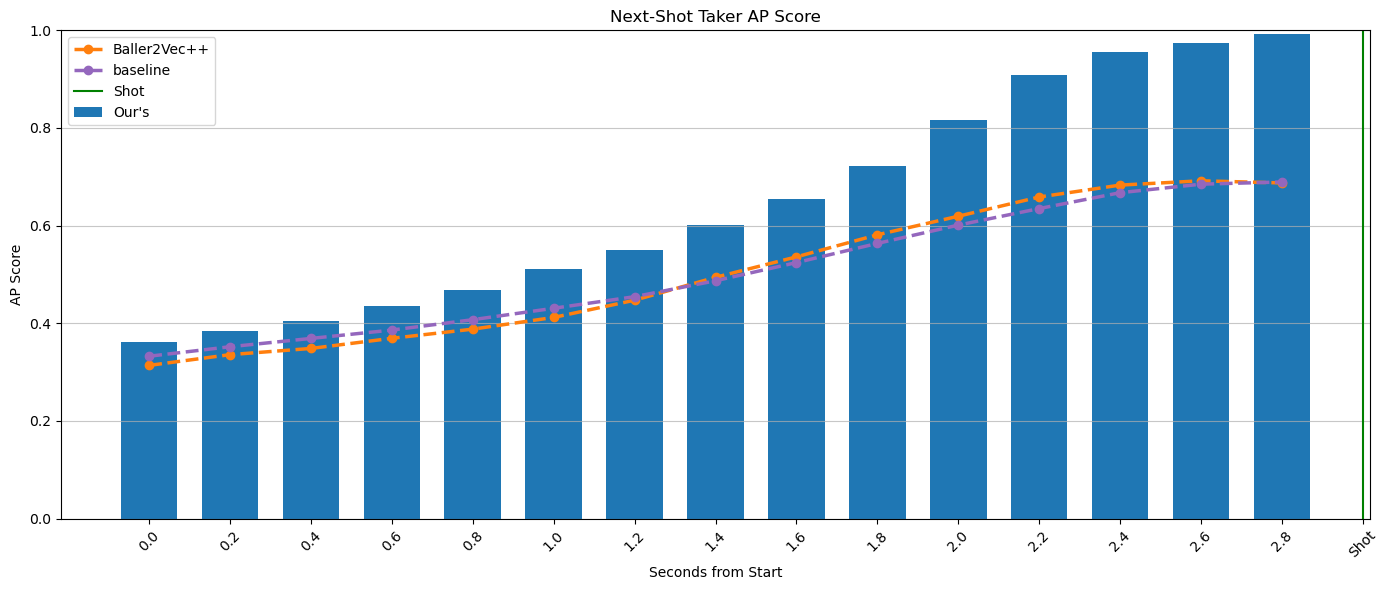}
    \caption{}
    \label{fig:shot_AP_timestep}
\end{subfigure}
\hfill
\begin{subfigure}[b]{0.48\textwidth}
    \includegraphics[width=\textwidth]{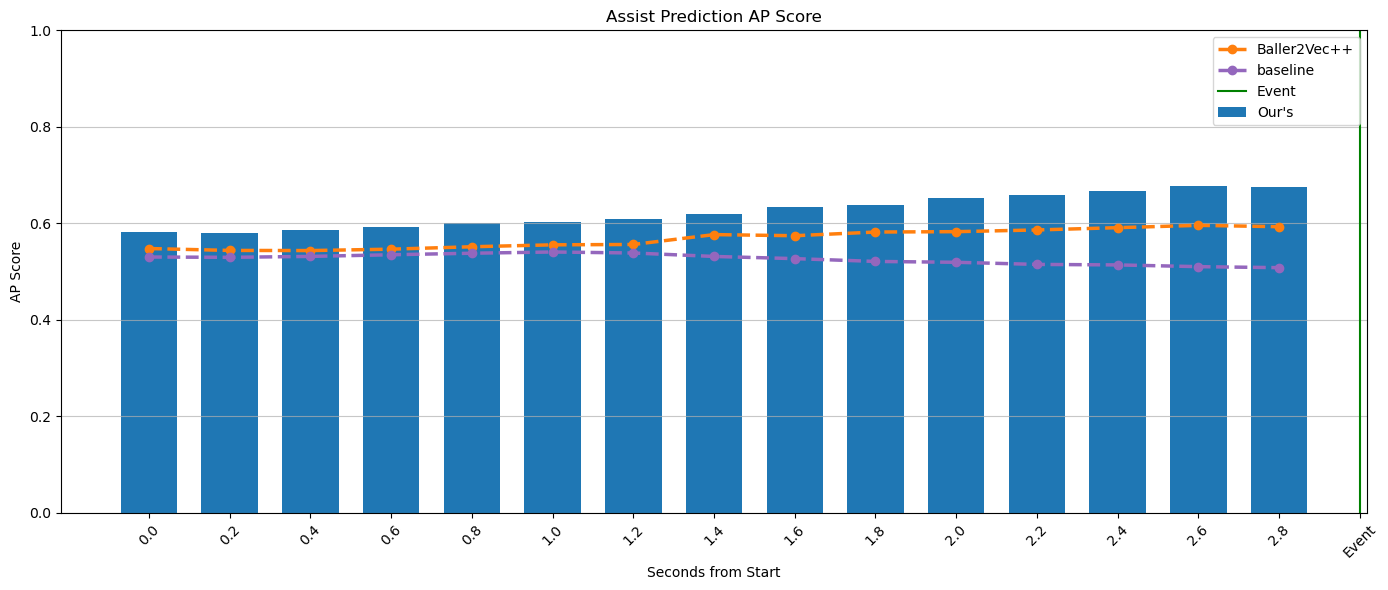}
    \caption{}
    \label{fig:assist_AP_timestep}
\end{subfigure}

\vspace{0.0cm}

\begin{subfigure}[b]{0.48\textwidth}
    \includegraphics[width=\textwidth]{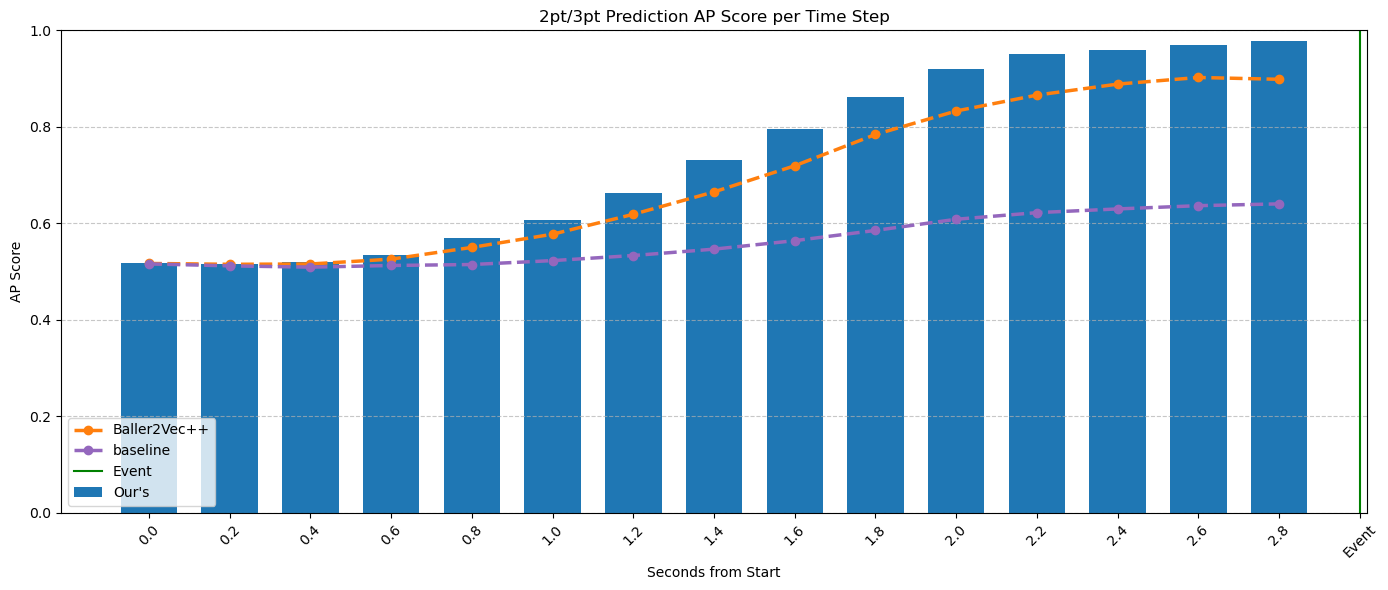}
    \caption{}
    \label{fig:2pt3pt_AP_timestep}
\end{subfigure}
\hfill
\begin{subfigure}[b]{0.48\textwidth}
    \includegraphics[width=\textwidth]{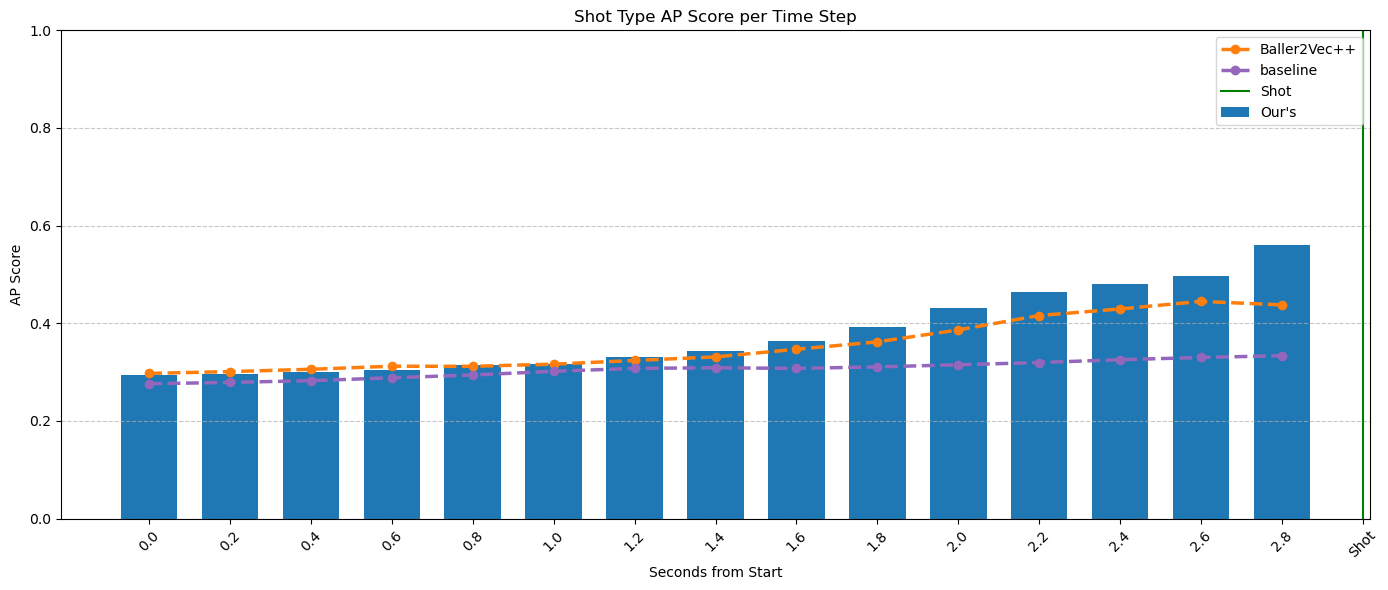}
    \caption{}
    \label{fig:shot_type_AP_timestep}
\end{subfigure}

\caption{AP scores across time horizons for downstream prediction tasks, showing consistent performance advantages of our approach over baselines as predictions approach the target event.}
\label{fig:all_downstream_AP}
\end{figure*}

As shown in Figure \ref{fig:shot_AP_timestep}, our model consistently outperforms both Baller2Vec++ and the baseline over all prediction horizons. At the early prediction point (2.8s before the shot), our model achieves 0.38 AP compared to Baller2Vec++'s 0.34 AP and the baseline's 0.35 AP. Although performance differences are modest at this early stage, reflecting the inherent difficulty of anticipating shots far in advance, a substantial performance divergence emerges as the shot approaches. Our model demonstrates a significantly steeper improvement curve, reaching an impressive 0.91 AP at 0.8s before the shot, compared to Baller2Vec++'s 0.66 AP and the baseline's 0.63 AP. This widening performance gap highlights our embedding's effectiveness at capturing the increasingly clear movement patterns that emerge as shooting actions develop. 

Figure \ref{fig:shot_embedding_comparison} reveals another important advantage of our approach: data efficiency. When training data is reduced exponentially, our model maintains relatively stable performance, while the baseline approach shows dramatic degradation. This superior data efficiency demonstrates how our two-stage pretraining approach effectively captures generalizable shooting patterns, allowing the model to make accurate predictions even with limited task-specific training examples.

\myparagraph{Pick Prediction Task}
The pick prediction task evaluates the ability to identify screening plays in basketball. We define a pick as a sequence where: (1) an offensive player with the ball is outside the 3-point line in the attacking half; (2) a teammate without the ball is stationary; (3) a defensive player is near both offensive players; and (4) within 1 second, the ball handler moves to the opposite side of the line connecting the stationary teammate and the basket. 


Our model outperforms both Baller2Vec++ and the baseline approach across all prediction horizons. At 0.6s before the pick, our model achieves 0.82 AP compared to Baller2Vec++'s 0.79 AP and the baseline's 0.70 AP, with the advantage maintained through to the moment of occurrence where we reach 0.85 AP. It should be noted that the maximum achieved AP (approximately 0.85) reflects the inherent complexity of precisely defining pick plays using geometric heuristics. Basketball screens exist on a spectrum of execution quality and effectiveness that can make categorical classification challenging even for expert analysts. Our heuristic-based detection captures the canonical form of picks, but naturally includes boundary cases where reasonable observers might disagree on classification (see Appendix~\ref{app:additional_results} for complete temporal analysis).

\myparagraph{Assist Prediction Task}
The assist prediction task evaluates a model's ability to anticipate passing plays leading directly to successful baskets. We define an assist as a sequence where: (1) an offensive player A passes the ball to teammate B; (2) teammate B takes no more than two dribbles after receiving the pass; (3) teammate B shoots and scores the basket within 3 seconds of receiving the pass. This challenging task requires understanding both passing intentions and predicting successful scoring outcomes. As Figure \ref{fig:assist_AP_timestep} shows, our model consistently outperforms both Baller2Vec++ and the baseline approach over all time horizons. The performance gap is evident even at early prediction points (2.8 seconds before the assist), where our model achieves 0.58 AP compared to Baller2Vec++'s 0.55 AP and the baseline's 0.53 AP. This advantage widens as the play develops, with our model reaching 0.65 AP at 0.8 seconds before completion, compared to Baller2Vec++'s 0.58 AP and the baseline's declining performance of 0.52 AP. This consistent detection advantage demonstrates our model's superior ability to recognize the coordinated movement patterns that precede successful assist plays. It is worth noting that even at the moment of the shot, our model reaches 0.65 AP rather than higher values. This ceiling reflects a fundamental characteristic of assist prediction: labels depend on shot success, which has inherent randomness due to minor variations in release angle, ball rotation, or rim bounce, limiting achievable performance for any motion-based model. 

\myparagraph{2-Point vs 3-Point Shot Prediction Task}
Predicting shot location- whether a play will result in a 2-point or 3-point attempt. At 1.6 seconds before the shot, our model achieves 0.73 AP compared to the baseline's 0.55 AP. This early advantage reflects our embedding's ability to capture court positioning patterns and movement tendencies that signal shot location intent. 

The performance gap widens dramatically as the shot approaches. At 0.4 seconds, our model reaches 0.97 AP versus the baseline's 0.64 AP. Figure \ref{fig:2pt3pt_AP_timestep} shows this progression across timesteps. The near-perfect late-stage performance demonstrates how skeletal features capture the distinctive body positioning and shooting mechanics that differentiate 2-point and 3-point attempts. 

\myparagraph{Shot Type Classification Task}
Shot type classification presents a harder challenge: distinguishing between dunks, hooks, jumpshots, and layups based on developing movement patterns. At 1.6 seconds before the shot, our model achieves 0.34 AP,  the baseline (0.31 AP). The modest absolute values reflect genuine difficulty in this task- many shot types remain ambiguous until late in their execution, and even human observers can struggle with borderline cases between similar mechanics.

Performance improves as the action unfolds, with our model reaching 0.50 AP at 0.4 seconds compared to the baseline's matching 0.33 AP. As shown in Figure \ref{fig:shot_type_AP_timestep}, the performance curves reveal how shot type becomes progressively clearer.

\subsection{Embedding Analysis and Ablation Study}
\begin{table*}[t]
\centering
\caption{Performance comparison across tasks and time horizons (Average Precision)}
\label{tab:ablation_analysis}
\resizebox{\textwidth}{!}{%
\begin{tabular}{l|cc|cc|cc|cc|cc}
\toprule
\multirow{2}{*}{Embedding} & \multicolumn{2}{c|}{Pick} & \multicolumn{2}{c|}{Shot Taker} & \multicolumn{2}{c|}{Assist} & \multicolumn{2}{c|}{2pt/3pt} & \multicolumn{2}{c}{Shot Type} \\
\cmidrule(lr){2-3} \cmidrule(lr){4-5} \cmidrule(lr){6-7} \cmidrule(lr){8-9} \cmidrule(lr){10-11}
& 0.6s & 0.0s & 2.8s & 0.8s & 2.8s & 0.8s & 1.6s & 0.4s & 1.6s & 0.4s \\
\midrule
Ours & \textbf{0.82} & \textbf{0.85} & \textbf{0.38} & \textbf{0.91} & \textbf{0.58} & 0.65 & \textbf{0.73} & \textbf{0.97} & \textbf{0.35} & \textbf{0.50} \\
Ours w/o Shoulder Normal & \textbf{0.82} & \textbf{0.85} & \textbf{0.38} & 0.90 & 0.57 & \textbf{0.67} & \textbf{0.73} & \textbf{0.97} & \textbf{0.35} & 0.48 \\
Ours w/o GNN & \textbf{0.82} & \textbf{0.85} & 0.36 & 0.77 & 0.55 & 0.61 & 0.67 & 0.93 & 0.32 & 0.42 \\
Ours w/o Events Projection & 0.78 & 0.83 & 0.33 & 0.80 & 0.55 & 0.56 & 0.70 & \textbf{0.97} & \textbf{0.35} & 0.47 \\
Ours w/o GNN w/o Events Projection& 0.79 & 0.82 & 0.33 & 0.65 & 0.55 & 0.56 & 0.67 & 0.91 & 0.34 & 0.42 \\
Ours w/o Shoulder Normal w/o Events Projection& 0.78 & 0.81 & 0.33 & 0.78 & 0.56 & 0.58 & 0.70 & \textbf{0.97} & 0.33 & 0.45 \\
\midrule
Baller2Vec++ w/ Events Projection& 0.80 & 0.82 & 0.36 & 0.77 & 0.56 & 0.61 & 0.67 & 0.93 & 0.33 & 0.43 \\
Baller2Vec++ & 0.79 & 0.82 & 0.34 & 0.66 & 0.55 & 0.58 & 0.67 & 0.90 & 0.33 & 0.44 \\
\midrule
TranSPORTmer & 0.71 & 0.72 & 0.25 & 0.30 & 0.52 & 0.52 & 0.53 & 0.68 & 0.29 & 0.33 \\
Baseline & 0.70 & 0.72 & 0.35 & 0.63 & 0.53 & 0.52 & 0.55 & 0.64 & 0.31 & 0.33 \\
\bottomrule
\end{tabular}
}
\end{table*}


Our ablation studies reported in Table~\ref{tab:ablation_analysis} provide valuable insights into the contribution of each model component across five challenging basketball analytics tasks. Full precision recall curves for additional time horizons are available in appendix \ref{app:additional_results}. 

\myparagraph{Impact of Skeletal Motion Data} Comparing ``Ours w/o Events Projection'' with Baller2Vec++ isolates the contribution of skeletal motion data. The skeletal embedding provides notable improvements, particularly for shot-taker prediction at later stages (0.80 vs 0.66 AP). This confirms our hypothesis that detailed motion information provides valuable cues about player intent. 

\myparagraph{Value of Event Projection Pretraining}
Comparing ``Ours'' to ``Ours w/o Events Projection'' reveals substantial benefits from multi-task pretraining. The event projection heads consistently improve performance across all tasks, with remarkable gains in shot taker prediction (0.91 vs 0.80 AP at late stage) and significant improvements for assist prediction (0.65 vs 0.56 AP at late stage). This suggests that explicitly modeling basketball events helps models learn representations that better capture the semantic meaning of movements. 

\myparagraph{Practical Implementation with Shoulder Normals}
For pick prediction, our ``Ours w/o GNN'' variant (which retains shoulder normal vectors and event projection) performs equally well as our full model (0.82 AP in the early stage, 0.85 AP at late stage). This represents a practical implementation that captures essential orientation information without requiring full skeletal tracking. 

\myparagraph{Two-Stage vs. Direct Prediction}
The baseline approach significantly underperforms our embedding-based methods in all tasks and time horizons. Despite accessing the same raw data, the direct prediction approach fails to match even the comparable Baller2Vec++ embedding model. This validates our two-stage methodology of pretraining rich embeddings followed by task-specific linear classification, demonstrating that our approach effectively distills complex patterns into transferable representations that generalize better to downstream tasks. 

\myparagraph{Comparison with TranSPORTmer} TranSPORTmer \citep{capellera2024transportmer} represents recent work in multi-agent sports trajectory modeling. Our approach demonstrates consistent improvements across all five tasks, with substantial gains on shot taker prediction (0.91 vs 0.30 AP), shot location prediction (0.97 vs 0.68 AP), and shot type classification (0.50 vs 0.33 AP). These results validate that skeletal motion features capture crucial signals about player intent and shooting mechanics, while our event projection pretraining creates representations that encode both motion patterns and their strategic significance.

%% file: cvpr/sec/5_conclusion.tex
\section{Conclusion}
\label{sec:conclusion}
CourtMotion advances predictive analytics for basketball by connecting physical motion patterns with their tactical significance in the game context. Our experiments demonstrate a 35\% reduction in trajectory prediction error compared to position-only baselines, enabling reliable prediction of key game events seconds before they occur. This predictive capability represents a substantial improvement in multi-agent spatiotemporal modeling, particularly for environments with complex coordinated behaviors.
The consistent performance gains across all different tasks demonstrate the generalizability of our approach. Unlike end-to-end neural networks that struggle to capture the nuanced relationships between motion and intent, our hybrid GNN-Transformer architecture with event projection shows substantial advantages in both data efficiency and prediction accuracy.
Future research directions include expanding the event vocabulary to encompass more nuanced basketball actions, applying our framework to other team sports with similar coordinated dynamics, and developing specialized models for specific aspects of gameplay analysis. By integrating motion patterns with tactical outcomes, our approach provides both technical performance improvements and new analytical capabilities that enhance understanding of the complex, adversarial multi-agent behaviors in team sports.

%% file: cvpr/sec/6_appendix.tex
\section{Additional Experimental Results}
\label{sec:additional_results}
\label{app:additional_results}
\begin{figure*}[htbp]
 \centering
 \includegraphics[width=0.7\linewidth]{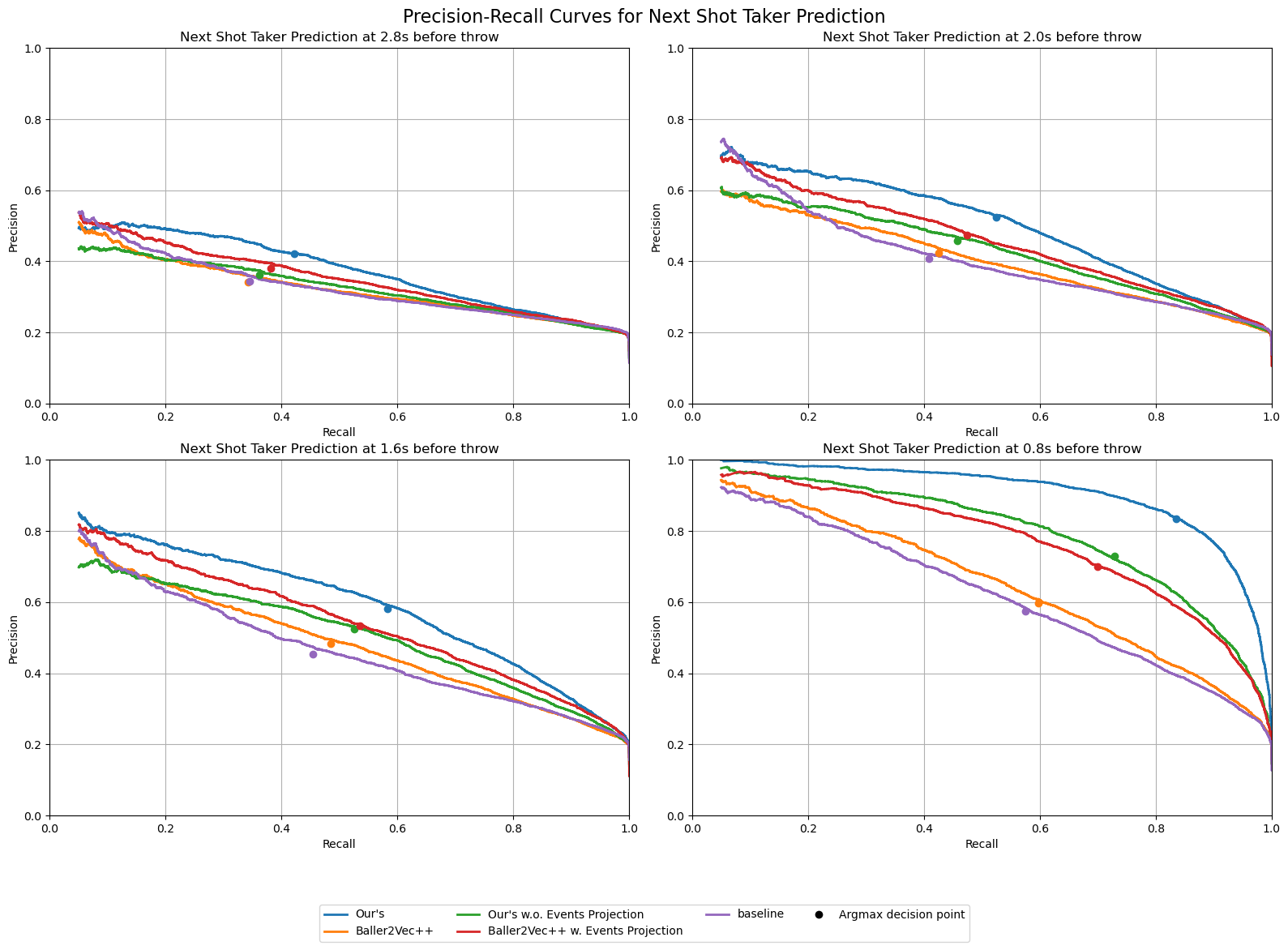}
 \caption{Precision-recall curves for shot taker prediction at four time horizons (2.8s, 2.0s, 1.6s, and 0.8s before shot).}
 \label{fig:app_shot_pr_curves}
\end{figure*}

\begin{figure*}[htbp]
 \centering
 \includegraphics[width=0.7\linewidth]{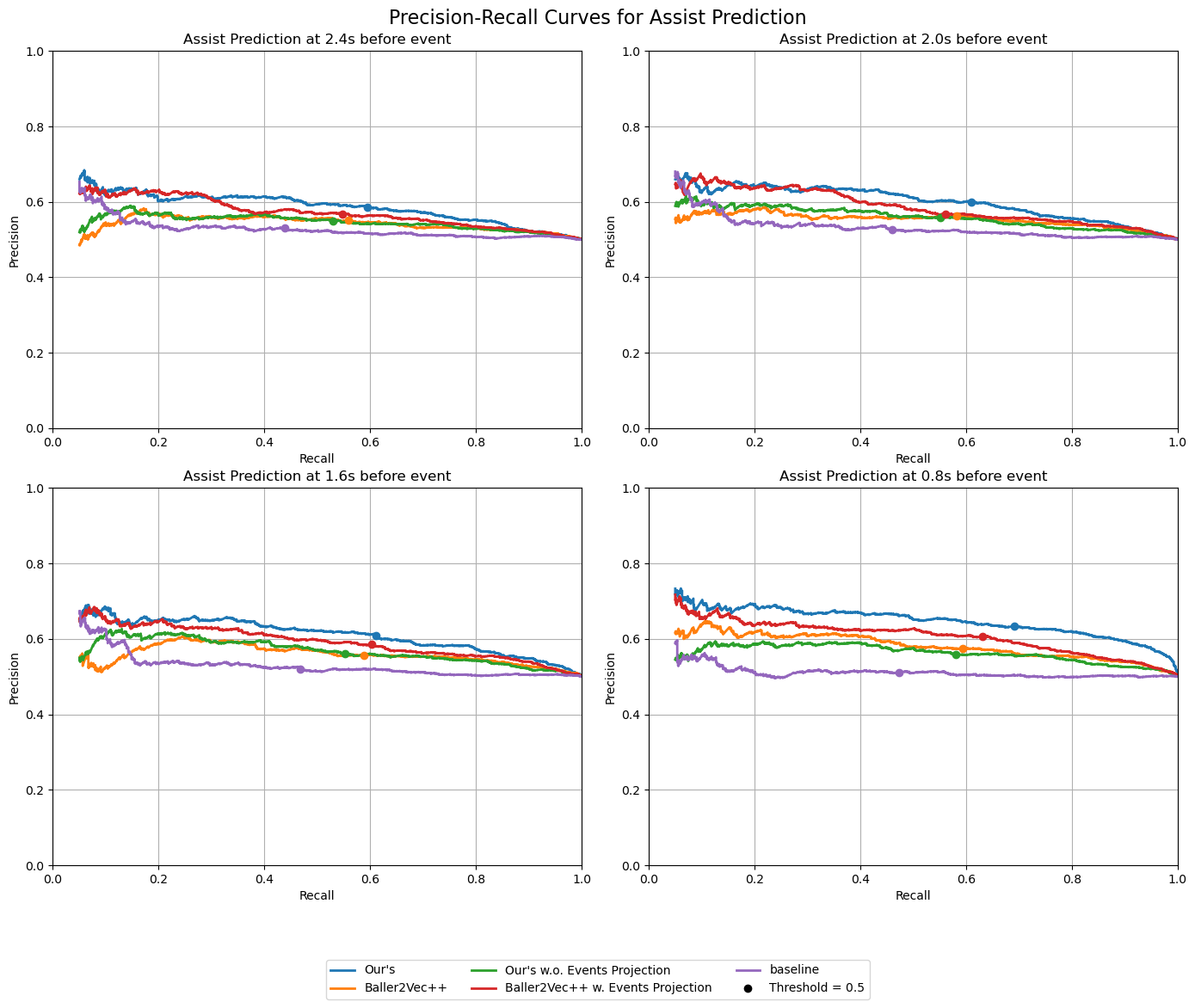}
 \caption{Precision-recall curves for assist prediction at 2.4s, 2.0s, 1.6s, 0.8s time horizons.}
 \label{fig:app_assist_pr_curves}
\end{figure*}

\begin{figure*}[htbp]
 \centering
 \includegraphics[width=0.7\linewidth]{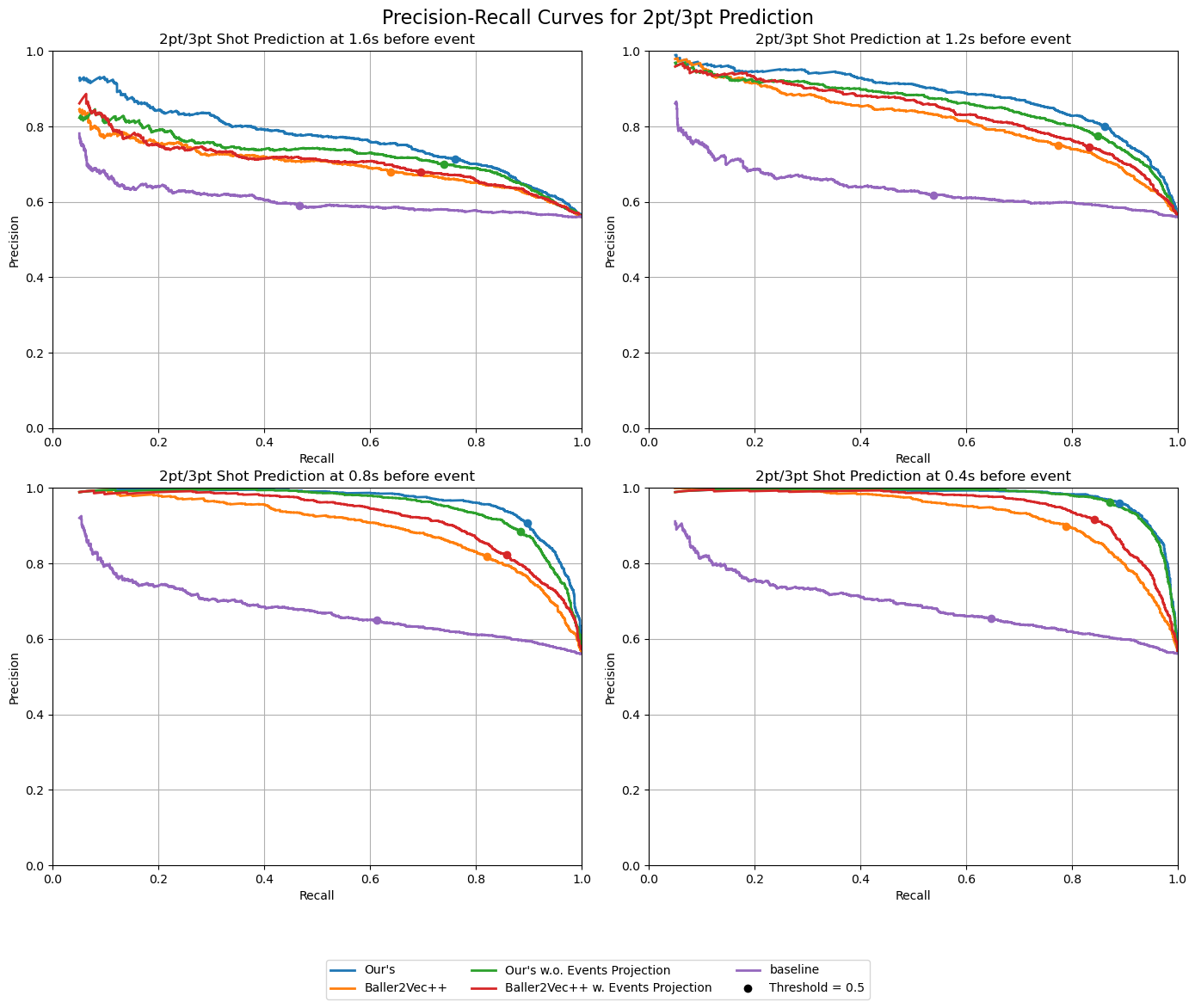}
 \caption{Precision-recall curves for 2pt/3pt shot location prediction at 1.6s, 1.2s, 0.8s, 0.4s time horizons.}
 \label{fig:app_2pt_3pt_pr_curves}
\end{figure*}

\begin{figure*}[htbp]
 \centering
 \includegraphics[width=0.7\linewidth]{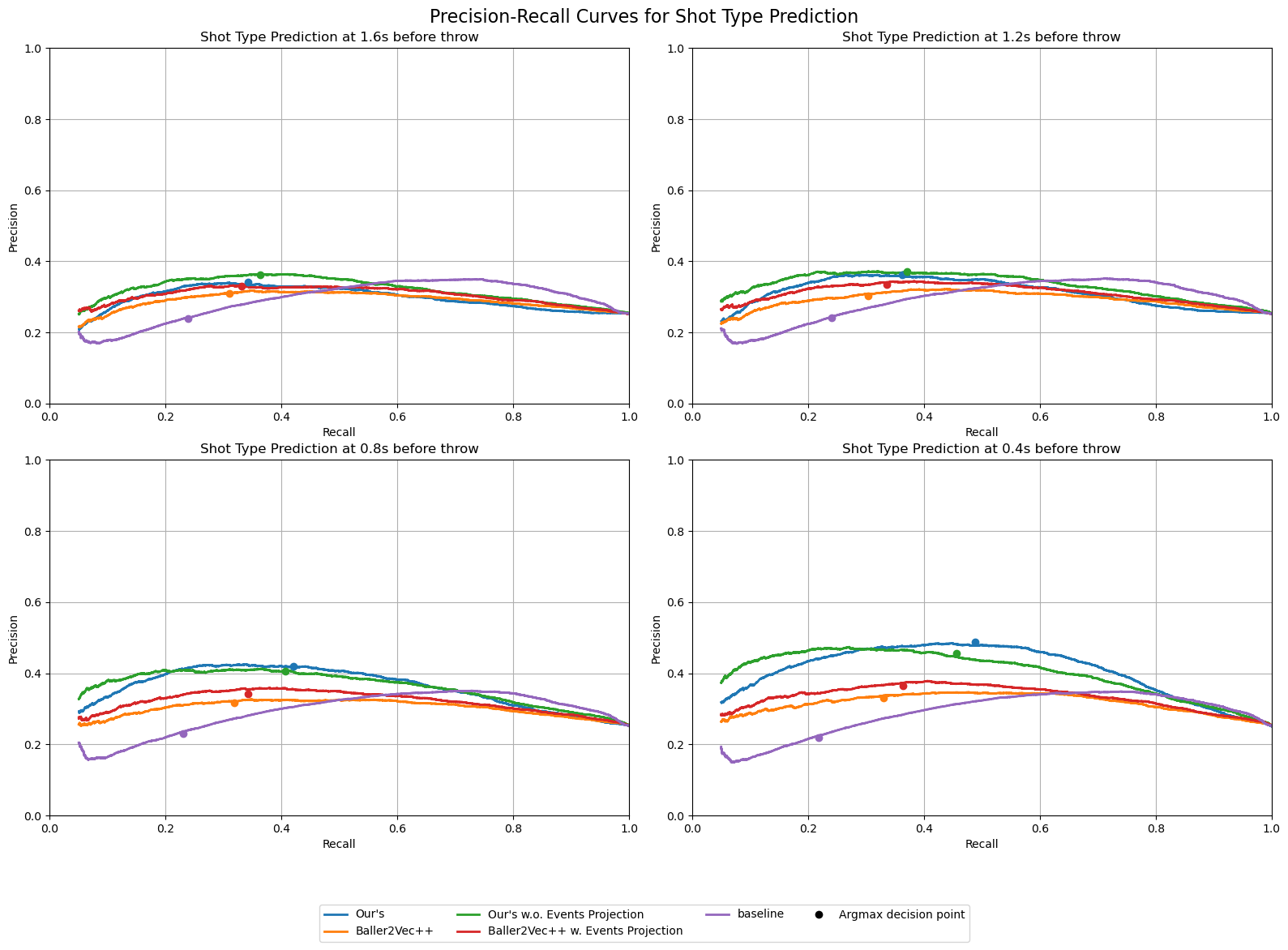}
 \caption{Precision-recall curves for shot type (dunk, hook, jump-shot, layup) prediction at 1.6s, 1.2s, 0.8s, 0.4s time horizons.}
 \label{fig:app_shot_type_pr_curves}
\end{figure*}

\begin{figure*}[htbp]
\centering
\begin{subfigure}[b]{0.5\linewidth}
    \includegraphics[width=\textwidth]{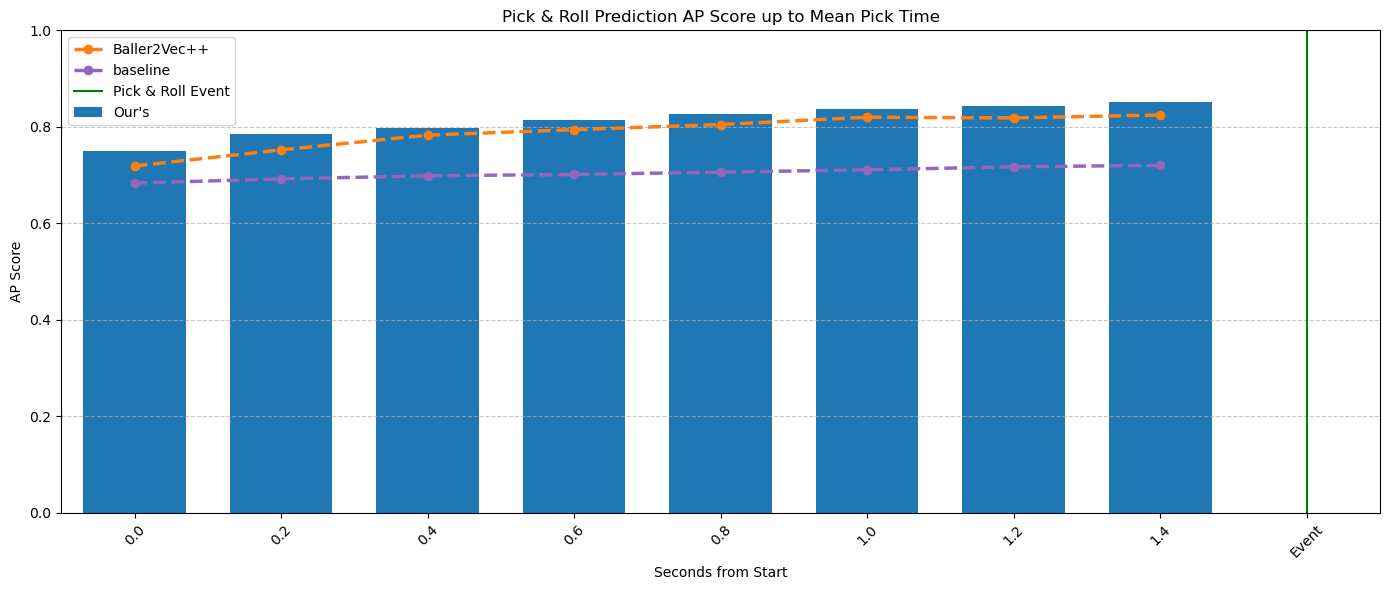}
    \caption{AP scores across time horizons}
    \label{fig:pick_AP_timestep}
\end{subfigure}

\vspace{0.3cm}

\begin{subfigure}[b]{0.7\linewidth}
    \includegraphics[width=\textwidth]{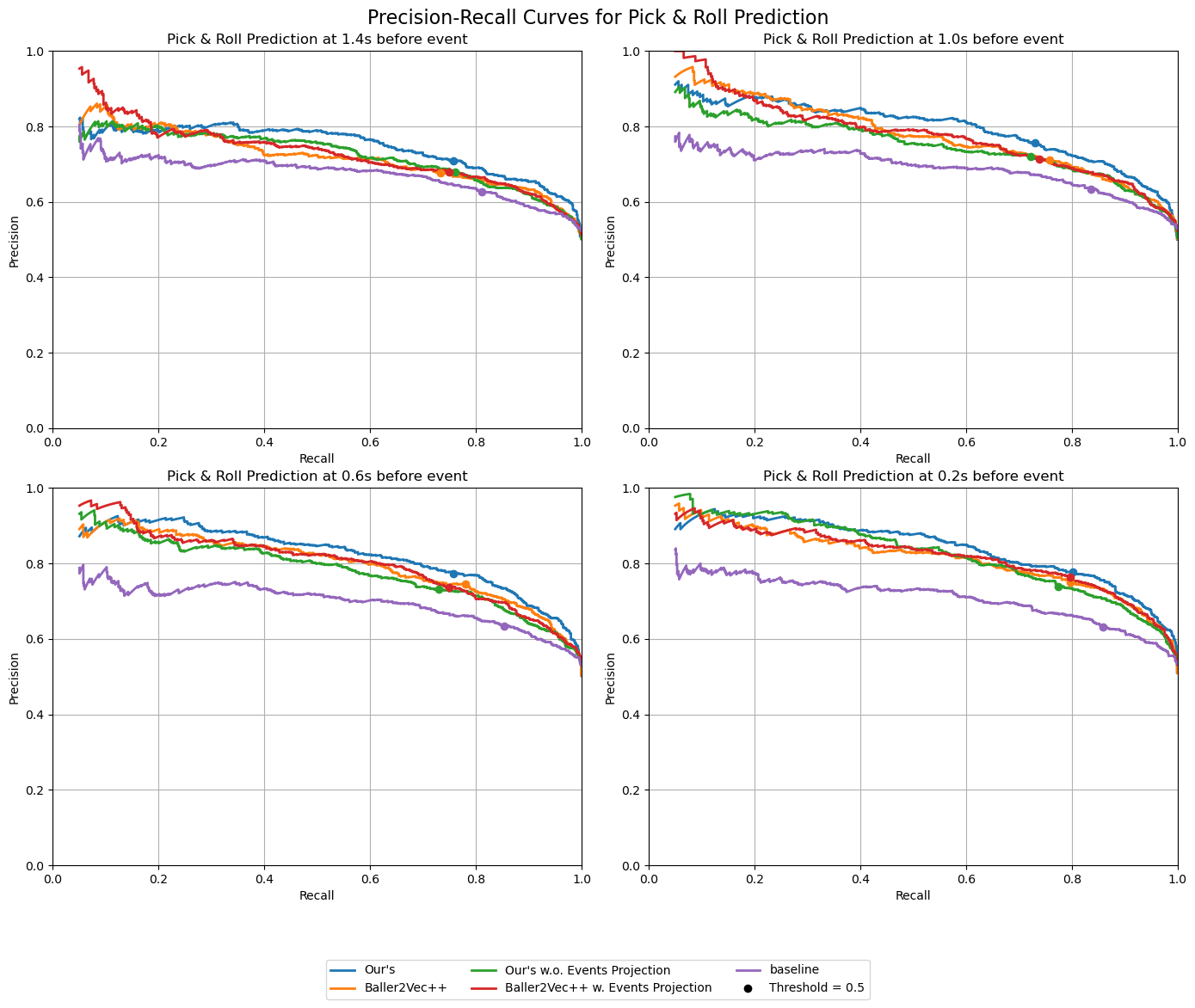}
    \caption{Precision-recall curves at multiple horizons}
    \label{fig:app_pick_pr_curve}
\end{subfigure}
\caption{Pick prediction results showing (a) AP scores from 1.4s before the pick to the moment of occurrence, and (b) precision-recall curves at 1.4s, 1.0s, 0.6s, 0.2s time horizons.}
\label{fig:app_pick_results}
\end{figure*}

This appendix provides additional experimental results for our downstream tasks. Precision-recall curves at various time horizons are shown in: Figure \ref{fig:app_shot_pr_curves} for shot taker prediction, Figure \ref{fig:app_assist_pr_curves} for assist prediction, Figure \ref{fig:app_2pt_3pt_pr_curves} for 2pt/3pt shot location prediction, Figure \ref{fig:app_shot_type_pr_curves} for shot type prediction, and Figure \ref{fig:app_pick_results} for pick prediction (including AP scores across time horizons).

